\documentclass[sigconf]{aamas}  % do not change this line!

\AtBeginDocument{%
  \providecommand\BibTeX{{%
    \normalfont B\kern-0.5em{\scshape i\kern-0.25em b}\kern-0.8em\TeX}}}
    
\usepackage{flushend} % do not change this line!
\usepackage[utf8]{inputenc} % allow utf-8 input
\usepackage{url}            % simple URL typesetting
\usepackage{booktabs}       % professional-quality tables
\usepackage{amsfonts}       % blackboard math symbols
\usepackage{nicefrac}       % compact symbols for 1/2, etc.
\usepackage{microtype}      % microtypography
\usepackage{color}      % microtypography
\usepackage{multirow}

\usepackage{comment}
\usepackage{float}
\usepackage{bm,amsmath}
\usepackage[inline]{enumitem}
\usepackage{subcaption}
\usepackage{mathtools}
\usepackage{makecell}
\usepackage[nopar]{lipsum}

\setcopyright{ifaamas}  % do not change this line!
\copyrightyear{2020} % do not change this line!
\acmYear{2020} % do not change this line!
\acmDOI{} % do not change this line!
\acmPrice{} % do not change this line!
\acmISBN{} % do not change this line!
\acmConference[AAMAS'20]{Proc.\@ of the 19th International Conference on Autonomous Agents and Multiagent Systems (AAMAS 2020)}{May 9--13, 2020}{Auckland, New Zealand}{B.~An, N.~Yorke-Smith, A.~El~Fallah~Seghrouchni, G.~Sukthankar (eds.)}  % do not change this line!

\def\Re{\mathbb{R}}

\def\Nat{{\rm I\kern\pIR N}}

\newcommand{\EE}[1]{\exptE\left[#1\right]}

\def\A{{\mathcal{A}}}

\def\D{{\mathcal{D}}}

\def\S{{\mathcal{S}}}

\def\vec0{{\boldsymbol{0}}}

\def\vecw{{\boldsymbol{w}}}

\newcommand{\ra}{\rightarrow}
\newcommand{\beq}{\begin{equation}}
\newcommand{\eeq}{\end{equation}}
\newcommand{\beqa}{\begin{eqnarray}}
\newcommand{\eeqa}{\end{eqnarray}}
\newcommand{\beqan}{\begin{eqnarray*}}
\newcommand{\eeqan}{\end{eqnarray*}}
\newcommand{\ben}{\begin{eqnarray*}}
\newcommand{\een}{\end{eqnarray*}}

\def\tr{^\top\!}

\renewcommand{\EE}[2]{\mathbb{E}_{#1\!\!}\left[#2\right]}

\newcommand{\CEE}[3]{\EE{#1}{{#2}~\middle\vert~{#3}}}
\renewcommand{\CEE}[3]{\EE{#1}{{#2}\mid{#3}}}

\def\CEpi#1#2{\CEE{\pi}{#1}{#2}}

\def\Epi#1{\EE{\pi}{#1}}

\begin{document}

\title{Improving Performance in Reinforcement Learning by \\ Breaking Generalization in Neural Networks}  % put your title here!
% \titlenote{Titlenote --- e.g., used for things like "This article extends an earlier paper titled XYZ", and "equal contribution by the first two authors".}

% AAMAS: as appropriate, uncomment one subtitle line; see camera ready instructions
%\subtitle{Extended Abstract}
%\subtitle{Blue Sky Ideas Track}
%\subtitle{JAAMAS Track}
%\subtitle{Doctoral Consortium}                              
%\subtitle{Demonstration}
%\subtitlenote{Please refrain from using subtitle notes}

% replace this with your author block!
\author{Sina Ghiassian, Banafsheh Rafiee, Yat Long Lo, Adam White}
\affiliation{%
 \institution{Reinforcement Learning and Artificial Intelligence Laboratory, University of Alberta and \\
 Alberta Machine Intelligence Institute (AMII)}
 \city{Edmonton, AB} 
 \state{Canada} 
}
\email{{ghiassia, rafiee, loyat, amw8}@ualberta.ca}


\begin{abstract}
Reinforcement learning systems require good representations to work well. For decades practical success in reinforcement learning was limited to small domains. Deep reinforcement learning systems, on the other hand, are scalable, not dependent on domain specific prior knowledge and have been successfully used to play Atari, in 3D navigation from pixels, and to control high degree of freedom robots. Unfortunately, the performance of deep reinforcement learning systems is sensitive to hyper-parameter settings and architecture choices. Even well tuned systems exhibit significant instability both within a trial and across experiment replications. In practice, significant expertise and trial and error are usually required to achieve good performance. One potential source of the problem is known as catastrophic interference: when later training decreases performance by overriding previous learning. Interestingly, the powerful generalization that makes Neural Networks (NN) so effective in batch supervised learning might explain the challenges when applying them in reinforcement learning tasks. In this paper, we explore how online NN training and interference interact in reinforcement learning. We find that simply re-mapping the input observations to a high-dimensional space improves learning speed and parameter sensitivity. We also show this preprocessing reduces interference in prediction tasks. More practically, we provide a simple approach to NN training that is easy to implement, and requires little additional computation. We demonstrate that our approach improves performance in both prediction and control with an extensive batch of experiments in classic control domains.
\end{abstract}

 % put your comma-separated keywords here!

\maketitle

%%%%%%%%%%%%%%%%%%%%%%%%%%%%%%%%%%%%%%%%%%%%%%%%%%%%%%%%%%%%%%%%%%%%%%%%%%%%%%%%%%%%%%%%%%%%%%%%%%%%%%%%%
%% start of main body of paper

\section{Introduction}
\label{sct:Introduction}

Reinforcement learning (RL) systems require good representations to work well. For decades practical success in RL was restricted to small domains---with the occasional exception such as Tesauro’s TD-Gammon (Tesauro, 1995). High-dimensional and continuous inputs require function approximation where the features must either be designed by a domain expert, constructed from an exhaustive partitioning schemes (e.g., Tile Coding), or learned from data. Expert features can work well (Sturtevant and White, 2007 and Silver, 2009), but depending on prior knowledge in this way limits scalability. Exhaustive partition strategies can be extended beyond small toy tasks (Stone and Sutton, 2001; Modayil et al., 2014; Rafiee et al., 2019), but ultimately do not scale either. Neural Networks (NN), on the other hand, are both scalable and not dependent on domain specific prior knowledge. Unfortunately, training NNs is typically slow, finicky, and not well suited for RL tasks where the training data is temporally correlated, non-stationary, and presented as an infinite stream of experience rather than a batch.

The practice of combining neural network function approximation and reinforcement learning has significantly improved. Deep reinforcement learning systems have been successfully deployed on visual tasks like Atari, 3D navigation, and video games (Mnih et al., 2015; Parisotto and Salakhutdinov, 2017; Vinyals et al., 2019). Deep RL systems can control high degree of freedom robots (Riedmiller et al., 2018), and learn in robot simulation domains directly from joint angles and velocities (Duan et al., 2016). All these systems rely on a combination of improved optimization algorithms (Kingma and Ba, 2014), Experience Replay (Lin, 1992), and other tricks such as Target Networks (Mnih et al., 2015).

There are many challenges in designing and training NN-based RL systems. Many systems exhibit extreme sensitivity to key hyper-parameters (Henderson et al., 2018)---choices of replay buffer size (Zhang and Sutton, 2017), optimizer hyper-parameters (Jacobsen et al., 2019), and other algorithm-dependent hyper-parameters have a large impact on performance. Many systems exhibit fast initial learning, followed by catastrophic collapse in performance, as the network unlearns its previously good policy (Goodrich, 2015). In some domains, simpler learning systems can match and surpass state-of-the-art NN-based alternatives (Rajeswaran et al., 2017).

Perhaps many of these frailties can be largely explained by aggressive generalization and interference that can occur in neural network training. The concept of Catastrophic Interference is simple to explain: training on a sequence of tasks causes the network to override the weights trained for earlier tasks. This problem is particularly acute in RL, because the agent's decision making policy changes over time causing interference to occur during single task training (Kirkpatrick et al., 2017; Liu et al., 2019). There are three primary strategies for dealing with interference: (1) adapting the loss to account for interference (Javed and White, 2019), (2) utilising networks that are robust to interference (Liu et al., 2019), or (3) adapting the training---classically by shuffling the training data in supervised learning (French, 1999), or via experience replay in RL. 

In this paper we propose a new approach to reducing interference and improving online learning performance of neural-network based RL. Our idea is based on a simple observation. Deep RL systems are more difficult to train in domains where there is significant and inappropriate generalization of the inputs. For example, in Mountain Car the observations are the position and velocity of the car. A NN will exploit the inherent generalization between states that are close in Euclidean space. However, the value function exhibits significant discontinuities and it is difficult for the network to overcome the harmful generalization to learn a good policy. This inappropriate generalization is prevalent in many classic RL control domains, and could compound the effects of interference, resulting in slow learning. Inappropriate generalization is less prevalent in visual tasks because standard architectures utilize convolutional layers which are designed to manage the input generalization. 

Our proposed solution is to simply map the observations to a higher-dimensional space. This approach significantly reduces the harmful generalization in the inputs, has low computational overhead, and is easily scaled to higher dimensions. In fact, this is an old idea: randomly projecting the inputs was a common preprocessing step in training perceptrons (Minsky and Papert, 2017) and can be competitive with networks learned via Backprop (Sutton and Whitehead, 1993; Mahmood and Sutton, 2013). We explore input preprocessing based on simple independent discretization, and Tile Coding (Sutton and Barto, 2018). We show that our input preprocessing improves the learning speed of several standard neural-network systems, and reduces interference. In fact, DQN (Mnih et al., 2015) achieves low interference and efficient learning, possibly because it uses experience replay and target networks which reduce interference, as our experiments suggest. Across the board, our results show our input preprocessing strategy never reduced performance, and in many cases dramatically improved learning speed and hyper-parameter sensitivity. Our results show that neural-network learners can be made more user friendly and exhibit reliable and efficient training.

\section{Background}
\label{sct:Background}
This paper studies both prediction---value function approximation---and control---maximizing reward. We employ the Markov Decision Process (MDP) framework (Puterman, 2014). In this framework, an agent and environment interact at discrete time steps $t=0, 1, 2, \ldots$. At each time step, $t$, the agent is in a state $S_t \in \S$ and takes an action $A_t \in \A$, where $\S$ and $\A$ are state and action spaces respectively. The agent takes actions according to a policy $\pi:\A\times\S\ra[0, 1]$. In response to the action, the environment emits a reward $R_{t+1} \in \Re$ and takes the agent to the next state $S_{t+1}$. The environment makes this transition according to transition function $P(S_{t+1} | S_t, A_t)$.

In prediction (policy evaluation), the policy $\pi$, is fixed. The goal is to estimate the value function, defined as the expected return ($G_t\in\mathbb{R}$) if the agent starts from a state $s$ and follows policy $\pi$ until termination: 
\begin{align*}
v_{\pi}(s) \doteq \CEpi{G_t}{S_t=s} \doteq \CEpi{\sum_{k=0}^{\infty} {\gamma^k R_{t+k+1}} }{S_t=s}, %\label{eq:state-value-def} \nonumber
\end{align*}
for all $s\in\S$, where $\Epi{.|.}$ denotes the conditional expectation of a random variable under $\pi$ and $\gamma\in[0,1]$ is a scalar discount factor parameter.

In the control setting, the policy is not fixed. The agent seeks to find a policy that maximizes the expected return. In control, state-action value functions replace state value functions from the policy evaluation case. The state-action value is defined as:
\begin{align*}
q_{\pi}(s, a) &\doteq \CEpi{G_t}{S_t=s, A_t=a} \\
&\doteq \CEpi{\sum_{k=0}^{\infty} {\gamma^k R_{t+k+1}} }{S_t=s, A_t=a} .
%\label{eq:action_value_def} \nonumber
\end{align*}

We consider the case in which the state space is large and we cannot estimate one value for each state or state-action pair. Instead, we seek to make use of a parametric function to approximate the value function. We denote the approximate value function for states and state-action pairs by $\hat{v}(s, \vecw)$ and $\hat{q}(s, a, \vecw)$ respectively, where $\hat{v}(s, \vecw)\approx v_{\pi}(s)$ and $\hat{q}(s, a, \vecw)\approx q_\pi(s, a)$ where vector $\vecw\in\mathbb{R}^d$ includes the learned parameters of the approximation.

To estimate the state values, we use temporal-difference learning, more specifically, TD(0) (Sutton, 1988), to update the neural network parameters at each time step, where the neural network is the function approximator to TD(0). Let $\vecw$ be the weights and $\alpha$ be a small step-size. The network parameters are updated according to:
$$ 
\vecw_{t+1} \leftarrow \vecw_t + \alpha \, \delta_t \nabla_{\vecw}\hat{v}(S_t, \vecw_t). %\label{eq:TD_update}
$$
$\nabla_{\vecw}\hat{v}(S_t, \vecw_t)$ denotes the gradient of the function $\hat{v}(S_t, \vecw_t)$ with respect to the parameters $\vecw$ where $\vecw = \vecw_t$.
$\delta_t$ is called the temporal-difference error: $\delta_t \doteq R_{t+1} + \gamma \hat{v}(S_{t+1}, \vecw_t) - \hat{v}(S_t, \vecw_t).$

To estimate the state-action pair values, we use TD(0)'s control variant, called Sarsa(0) (Rummery and Niranjan, 1994). Sarsa(0) update rules are the same as TD(0)
except that $\hat{v}(S_t, \vecw_t)$ is replaced by $\hat{q}(S_t, A_t, \vecw_t)$. Sarsa(0) typically uses an $\epsilon$-greedy policy with respect to the state-action values to select actions.

Neural networks have been widely used to approximate value functions in reinforcement learning tasks, but there are several additional ingredients that improve performance. Often, neural networks are combined with Experience Replay (ER) (Lin, 1992), Target Networks (TN), and step-size adaptation methods (called optimizers). ER is a way to reuse previous data stored in a replay buffer to increase sample efficiency. The idea is simple, the agent stores recent experience in a buffer of experience tuples---$(S_t, A_t, R_{t+1}, S_{t+1})$---and replays it repeatedly, as if the agent was re-experiencing the data. The experience is typically sampled randomly from the buffer to break the temporal dependencies in the training data. The main idea in using target networks is to stabilize the update used in TD methods. TD updates towards a bootstrap target: the update target contains the network's current estimate of the value function. Instead of changing the network in the target on every update, the {\em Target Network} is periodically set equal to the learned network. Optimizers such as Adam (Kingma and Ba, 2014) are used in place of a global learning rate in stochastic gradient descent (e.g., $\alpha$ above). Instead we use a vector of step-sizes---one for each weight---that change with time. These three ingredients were used in DQN (Mnih et al., 2014), which is perhaps one of the most widely used and robust deep reinforcement learning systems available today.

\section{Breaking Generalization in Neural Networks}
\label{sct:BreakingTheGeneralization}

Neural networks can forget what they learned in the past due to a phenomenon known as catastrophic interference. Interference happens when a neural network is trained on new data and it overwrites what it has learned in the past. This phenomenon can be related to neural network's global generalization. 

To alleviate the interference issue, we propose mapping the input to a higher dimensional space. Specifically, we propose discretizing or tile coding the input as a preprocessing step before feeding it to the neural network. This preprocessing step breaks the input generalization and our hypothesis is that it helps reduce the overly global generalization of neural networks and in turn reduces interference and improves performance. We propose two simple approaches for breaking the generalization in the input space as discussed below and test our hypothesis in later sections.

The first approach is to simply use binning to discretize each dimension of the input separately. In this case, each dimension of the input is covered with a one dimensional grid and a one-hot vector is created that has a one in the bin where the input lies in and zero everywhere else. The same is done for all of the input dimensions and then the resulting vectors are concatenated to create a long one-dimensional vector, which is the final representation fed to the neural network. We will simply refer to this method as discretization-neural network or the shorthand D-NN.

\begin{figure}[t]
      \centering
     \includegraphics[width=0.8\linewidth]{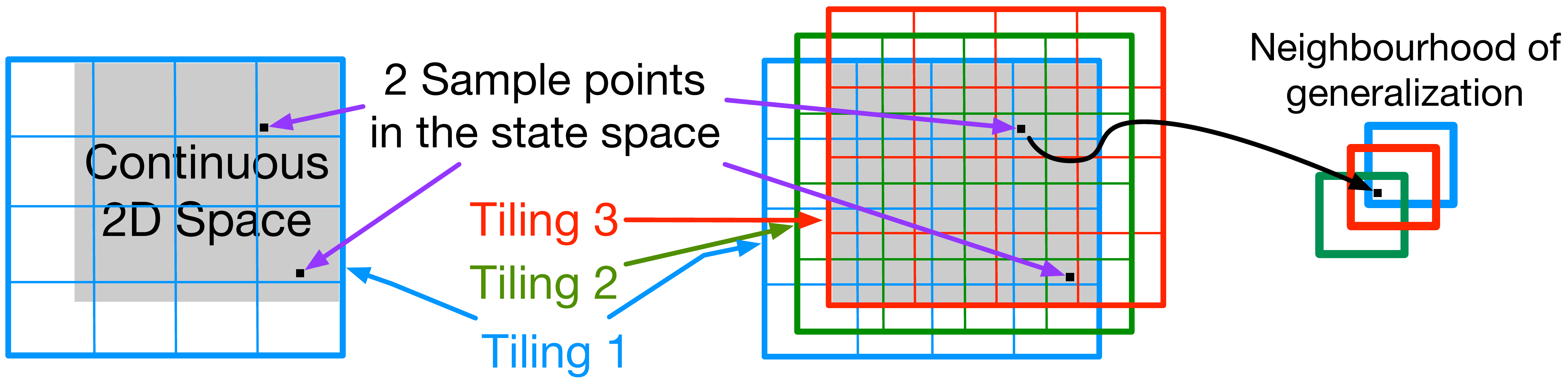}
      \caption{A continuous 2D space with 1 tiling on top of it is on the left. Three overlapping tilings on the 2D continuous space are shown in the middle in blue, green and red. The generalization region for a sample point is on the right.
      }
      \label{fig:tile_coding}
      \vspace{-0.5cm}
\end{figure}
The second approach that we use to break the generalization is Tile Coding (TC) (Albus 1975, 1981). We refer to this method with the shorthand TC-NN.
Tile coding works by covering the state space with a number of overlapping grids called tilings. Each grid divides the state space into small squares, called tiles. In Figure~\ref{fig:tile_coding}, a continuous 2D space is covered by 3 tilings where each tiling has 4 tile across each dimension (overall 16 tiles). Tile coding creates a representation for each point in space by concatenating the representation it has for each tiling. The representation for each tiling consists of a one hot vector that has a one for the tile that the point falls within and zero otherwise. For example, the representation for the point in Figure~\ref{fig:tile_coding} will have three ones in a vector of size 48 (3 tilings $\times$ $4\times 4$ tiles). See Sutton and Barto (2018) for a thorough explanation of tile coding.

Breaking the generalization in the input space increases the ability of the neural network to respond to different parts of the state space locally. With ReLU gates, the activation region of a node is the open half-space:  $\{ x \mid \langle w, x \rangle + b > 0 \}$ where $w$ represents the weights and $b$ is the bias associated with a node. If the NN receives raw observations as input, every node will respond to an entire half-space in the input space and might cause undesirable generalization. However, when the generalization is broken in the input space using discretization, each small area in the input space is mapped to a vertex of a hypercube. These vertices are all extreme points of a convex set and thus the ReLU activations will have the ability to respond to each of these sub-areas separately.

\begin{figure}[t]
      \centering
      \includegraphics[width=0.75\linewidth]{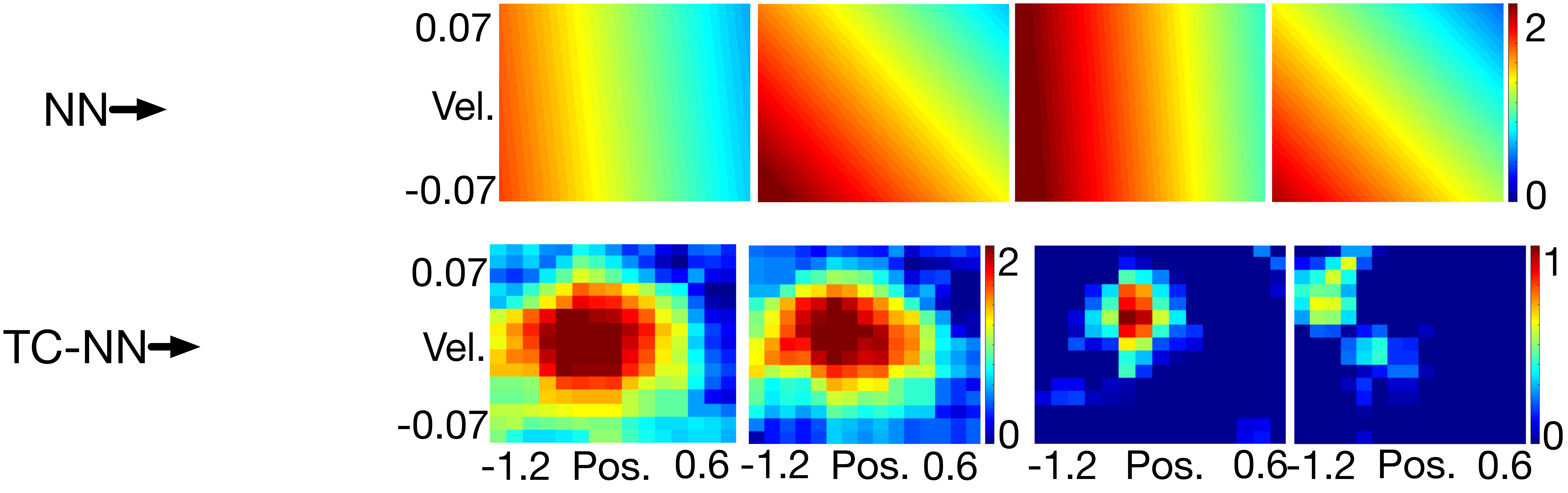}
      \caption{Response functions with raw inputs (top) and tile coding preprocessing (bottom) for Mountain Car control.}
      \label{fig:nn_and_tcnn_feature_maps}
\vspace{-0.5cm}
\end{figure}

Figure~\ref{fig:nn_and_tcnn_feature_maps} shows heat-maps for the case where NN used raw inputs (top) and tile coding preprocessing (bottom). The feature maps were created using a neural network trained on the Mountain Car problem for 500 episodes. Each heat-map represents the magnitude of the output of a node from the first hidden layer. Heat-maps on the bottom row of Figure~\ref{fig:nn_and_tcnn_feature_maps} show two rather global and two rather local node responses from the hidden layer. As shown in the figure, responses from the neural net that use raw inputs are global. So far we discussed the features of the proposed methods and have shown that the proposed method can have more local generalizations. However, what we have shown so far is more qualitative than quantitative. The next section uses quantitative measures of interference for comparison.

\section{Experimental setup}
\label{sct:ExperimentalSetup}
The experiments described in the following sections are rather extensive and have many components. At the highest level we investigate the impact of input preprocessing on a variety of base NN learning systems. Our experiments include both prediction and control, in two classic reinforcement learning control tasks. The following sections describe the simulation problems, and the base NN learning systems we used.

\subsection{Simulation problems}
\label{subsct:SimulationProblems}

We investigate three different problems, one prediction problem and two control problems. The control problems that we used are Mountain Car and Acrobot. The prediction problem also uses the Mountain Car testbed, but with a fixed given policy. 

Mountain Car simulates an underpowered car on the bottom of a hill that should pass the finish line on top of the hill (Moore, 1991). The problem has two dimensions: position and velocity. The position can vary between -1.2 and 0.6, and the velocity varies between -0.07 and 0.07. There are three actions in each state, throttle forward, throttle backward, and no throttle. The car starts around the bottom of the hill randomly in a point uniformly chosen between -0.4 and -0.6. The reward is -1 for each time step before the car passes the finish line at the top of the hill. When the position becomes larger than 0.5, the agent receives a reward of 0 and the episode terminates. The problem is episodic and not discounted ($\gamma=1$).

In the control version of the Mountain Car problem, the agent seeks to find a policy that ends the episode as fast as possible. In control, episodes were cut-off after 1000 steps. In prediction variant of the problem, we used a simple {\em energy pumping policy}. This policy chooses the action in agreement with the current velocity: left if velocity is negative, right otherwise.

The Acrobot (Sutton, 1996) is similar to a gymnast. Its goal is to swing its feet above a bar it is hanging from. The problem has four dimensions, two angles and two angular velocities (all real-valued). There are three discrete actions: positive torque, negative torque, and no torque. The reward is -1 on each time step before the Acrobot swings its feet above over the bar to terminate the episode. The task is undiscounted with $\gamma=1$. Episodes are cutoff after 500 steps. We used the Open AI Gym implementation of the Acrobot problem (Brockman et al., 2016).

\subsection{Methods}
\label{subsct:Methods}

\begin{figure*}[]
      \includegraphics[width=0.94\linewidth]{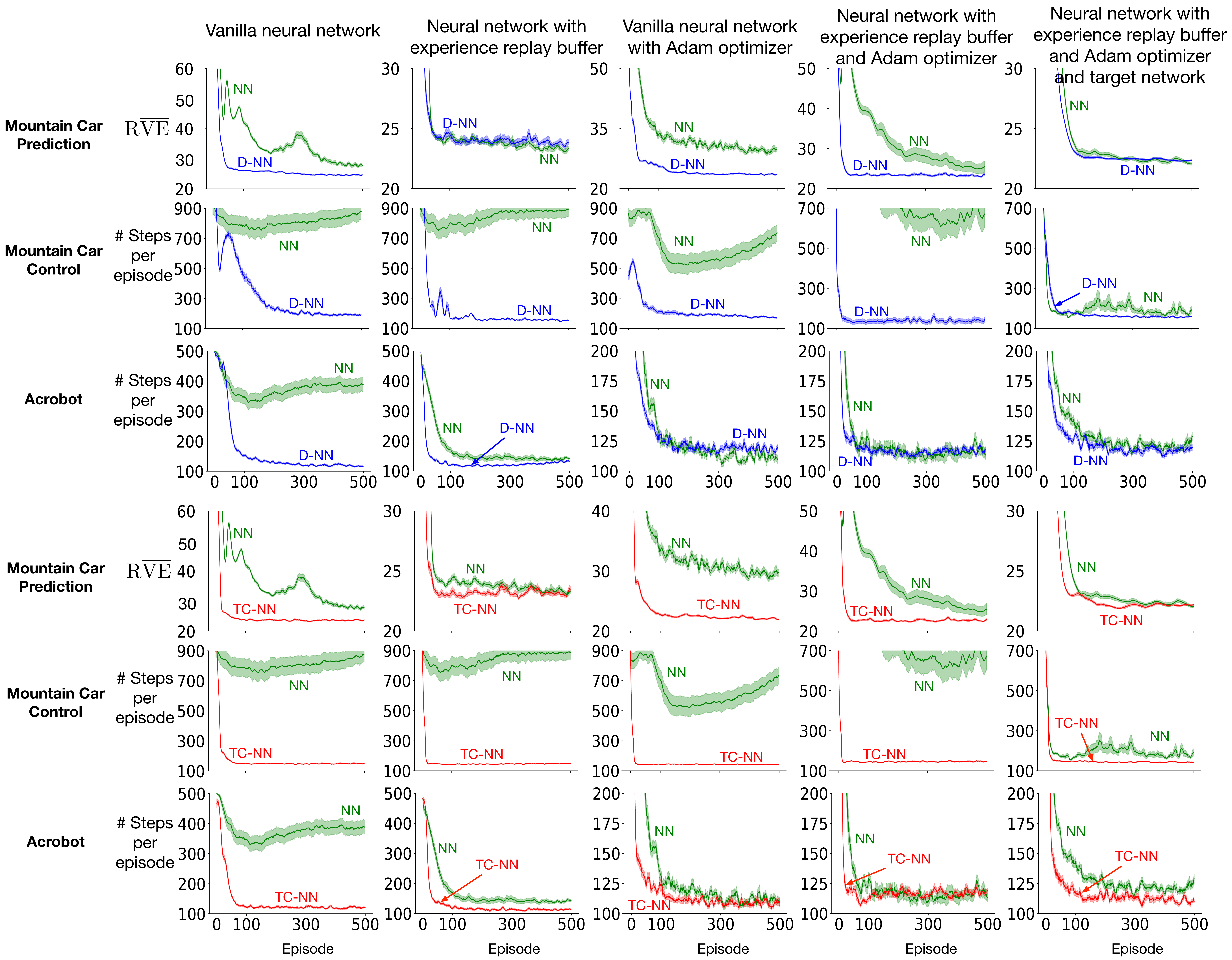}
      \caption{Learning curves for all tasks. Top three rows compare raw inputs with discretized inputs. Three bottom rows compare raw inputs with tile coding preprocessing. Discretizing/tile coding the input helped neural networks learn faster and converge to a better final performance. D-NN is short for Discretization+NN and TC-NN is short for Tile Coding+NN.}
      \label{fig:learning_curve}
\end{figure*}

\begin{figure*}[]
      \includegraphics[width=0.92\linewidth]{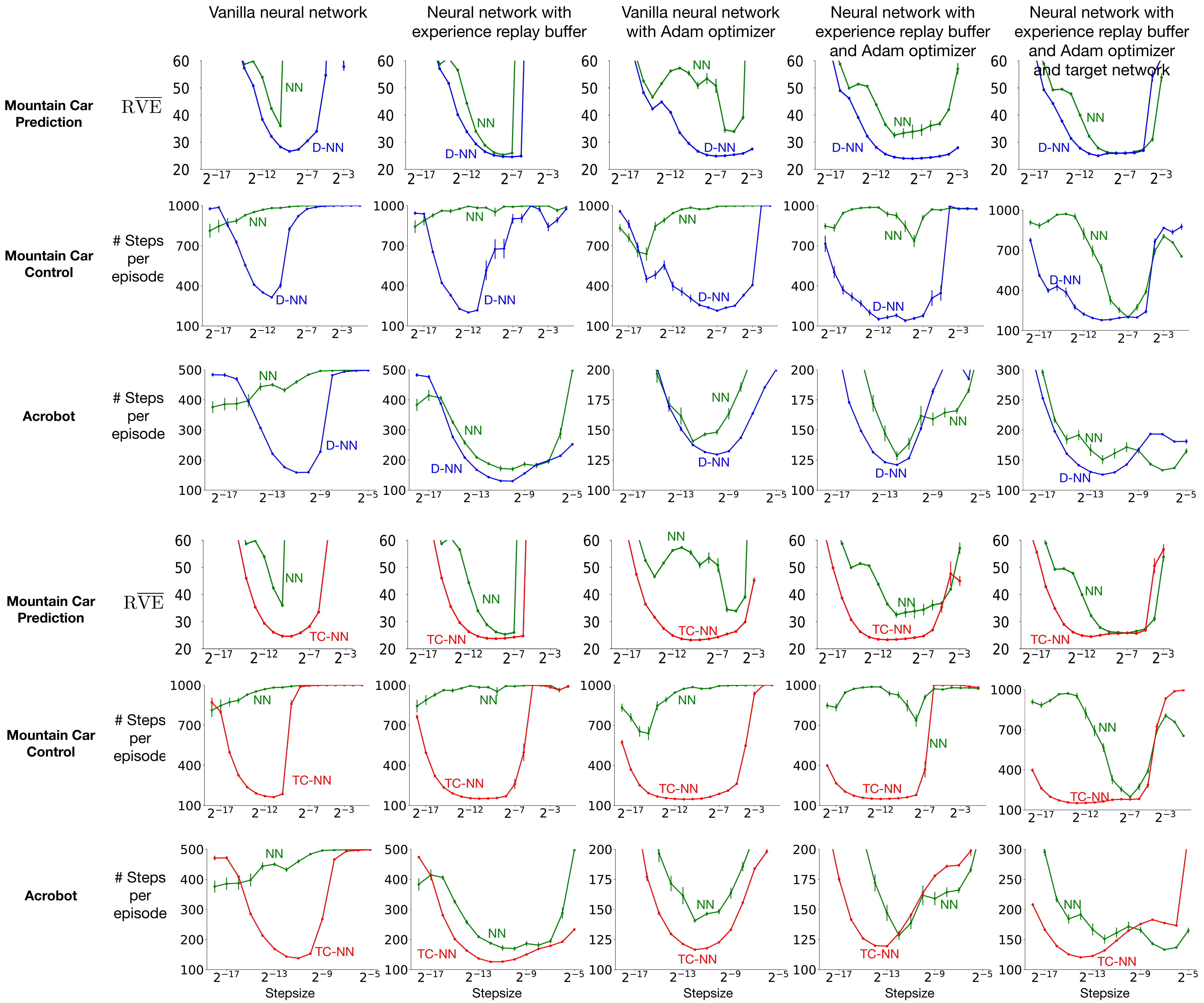}
      \caption{Step-size sensitivity curves over all tasks. Top three rows compare discretized and raw inputs. Three bottom rows compare tile coded preprocessing and raw inputs. Breaking the generalization in the input reduced sensitivity to step-size.}
      \label{fig:sensitivity}
\end{figure*}

We are interested in studying the impact of input preprocessing on a variety of NN learning systems. We investigate five NN learning systems that represent a spectrum from naive learning systems that we expect to work poorly, to well-known and widely used architectures. In particular we selected, (1) simple feed-forward NN with stochastic gradient descent (SGD); (2) simple feed-forward NN with ER and SGD; (3) simple feed-forward NN with Adam (Kingma and Ba, 2014); (4) simple feed-forward NN with ER, and Adam; (5) simple feed-forward NN with ER, Adam, and target networks.  

Given our five representative learning systems, we experiment wih three different input preprocessing strategies. The first involves no preprocessing, feeding in the raw input observations. The second, first discretizes the inputs before they are passed to the network. The third, first tile codes the observations. 

\section{Experimental Results}

Our experiments primarily focus on how different input preprocessing strategies impact both the speed of learning, and sensitivity to the step-size parameter values. In addition we also investigate how preprocessing impacts interference. To do so, we use a recently proposed interference measure (Liu, 2019). Our main hypothesis is that input preprocessing improves performance in prediction and control, because it reduces interference in the network updates. We begin first by reporting the improvements in prediction and control with discretization and tile coding. 

\subsection{Overall performance}
\label{subsct:OverallPerformance}

We used different measures to compare the algorithms for control and prediction tasks. In the control tasks, we simply report the number of steps it takes for the agent to finish the episode. Note that the reward at each time step for all tasks is -1 and thus the agent's objective is finish the task as fast as possible. 

For the Mountain Car prediction task, we used an easily interpretable error measure, the Root Mean Squared Value Error ($\text{R}\overline{\text{VE}}$). We measure, at the end of each episode, how far the learned approximatet value function is from the optimal value function. The ${\text{R}}\overline{\text{VE}}(\vecw_t)$ is defined as:
\begin{align}
\sqrt{\sum_{s\in\S}d_\pi(s)\left[\hat{v}(s, \vecw_t) - v_\pi(s)  \right]^2}
\approx\sqrt{ \frac{1}{|\mathcal{D}|} \sum_{s \in \mathcal{D}} \left[\hat{v}(s, \vecw_t) - v_{\pi}(s)\right]^2} \label{eq:RVE}
\end{align}
\noindent where $d_\pi(s)$ is the stationary distribution under $\pi$, $\hat{v}$ is the agent's estimate of the value function, and $v_{\pi}$ is the true value function. Since the state space is continuous, $d_\pi(s)$ is estimated by sampling states when following $d_\pi$. $\mathcal{D}$ is a set of states that is formed by following $\pi$ to termination and restarting the episode and following $\pi$ again. This was done for 10,000,000 steps, and we then sampled 500 states from the 10,000,000 states randomly. The true value $v_\pi(s)$ was simply calculated for each $s \in \D$ by following $\pi$ once to the end of the episode.

To create learning curves\footnote{All learning curves were smoothed using a sliding window of size 10.}, we ran each method with many different hyper-parameter combinations and plotted the one that minimized the Area Under the learning Curve (AUC)---total steps to goal in control and total $\text{R}\overline{\text{VE}}$ in prediction. We ran each method with each specific parameter setting 30 times (30 independent runs). We then averaged the results over runs and computed the standard error.

Figure~\ref{fig:learning_curve}, rows 1-3 compare each method using raw input--with its best performing hyper-parameters--with its counterpart that uses discretized inputs. In most cases, NNs with discretized inputs learned faster than NNs using raw inputs and converged to the same or a better final performance. Figure~\ref{fig:learning_curve}, rows 4-6 compares raw inputs with tile coding preprocessing. NNs with tile coding preprocessing outperforms NN with raw inputs.

In most cases, the difference between the performance of NN using remapped inputs and raw inputs were statistically significant (according to the standard error). The results suggest that preprocessing the inputs and projecting them into a higher dimensional space helps neural networks learn faster and more accurately. 

To further assess the significance of the difference in pairwise comparisons, we performed two sample t-tests. The pairs for the t-test were: 1) the method that used raw inputs \emph{and} 2) the method that used either tile coding or discretization preprocessing. Each one of the 30 plots in Figure~\ref{fig:learning_curve} includes two learning curves. Each plotted learning curve is the average over 30 learning curves of independent runs. We first averaged the performance (the value error for Mountain Car prediction, the number of steps per episode for Mountain Car control and Acrobot) over episodes for each run. This produced 30 numbers (one for each run) for each of the learning curves in each plot in Figure~\ref{fig:learning_curve}. We then used these 60 numbers (30 for each group) for a two sample t-test. Appendix A summarizes the p-values of the t-test in a table. The test revealed that in cases that tile coding or discretization improved performance, the improvement was statistically significant. %The significance test rejects the null hypothesis that the area under the learning curves come from normal distribution with equal means at the 5\% level.

\subsection{Sensitivity to step-size}
\label{subsct:SensitivityToStepsize}

In our previous experiment, we reported the results using the best performing step-size, which is not always feasible in practice. Ideally, an algorithm will perform well with many different settings of the step-size parameter. In this section, we investigate the performance of each of our five learning systems with many different step-size parameter values. Our objective is to understand how the input preprocessing interacts with step-size sensitivity. 

 We evaluated the performance using different step-sizes (see Table~\ref{tbl:parameter_list} for a list of step-sizes). Figure~\ref{fig:sensitivity} summarizes the sensitivity to step-size for the methods that used SGD and the initial step-size for the methods that used Adam. To create sensitivity curves, we ran each method with a specific step-size parameter value 30 times. We then averaged \emph{each run's} performance to get a single number that represents the area under the learning curve for that specific run for that specific method with the specific parameter settings. We then computed the average and standard error over the 30 numbers (one for each run) and plotted this number for each specific step-size in Figure~\ref{fig:sensitivity}. 

Each learning system that we tested actually has several hyper-parameters, and these can also have a significant impact on performance. For example, the algorithms that use the Adam optimizer have two extra parameters that specify the exponential decay rate for the first and second moment estimates, which are typically referred to as $\beta_1$ and $\beta_2$. To create a single curve for step-size sensitivity, we first searched over all the parameters to find the minimum AUC, we then fixed $\beta_1$ and $\beta_2$ to the values that achieved the minimum AUC and plotted the AUC over the step-size for those values.

As we see in Figure~\ref{fig:sensitivity}, discretizing/tile coding reduced sensitivity to step-size (compare the size of the bowl-shaped curves when using tile coding or discretization to raw inputs). The standard errors in Figure~\ref{fig:sensitivity} (some of which are not visible due to being small) show that the difference in the errors are statistically significant.

\subsection{Interference}
\label{subsct:InterferenceOverTime}

\begin{figure*}[]
      \includegraphics[width=0.8\linewidth]{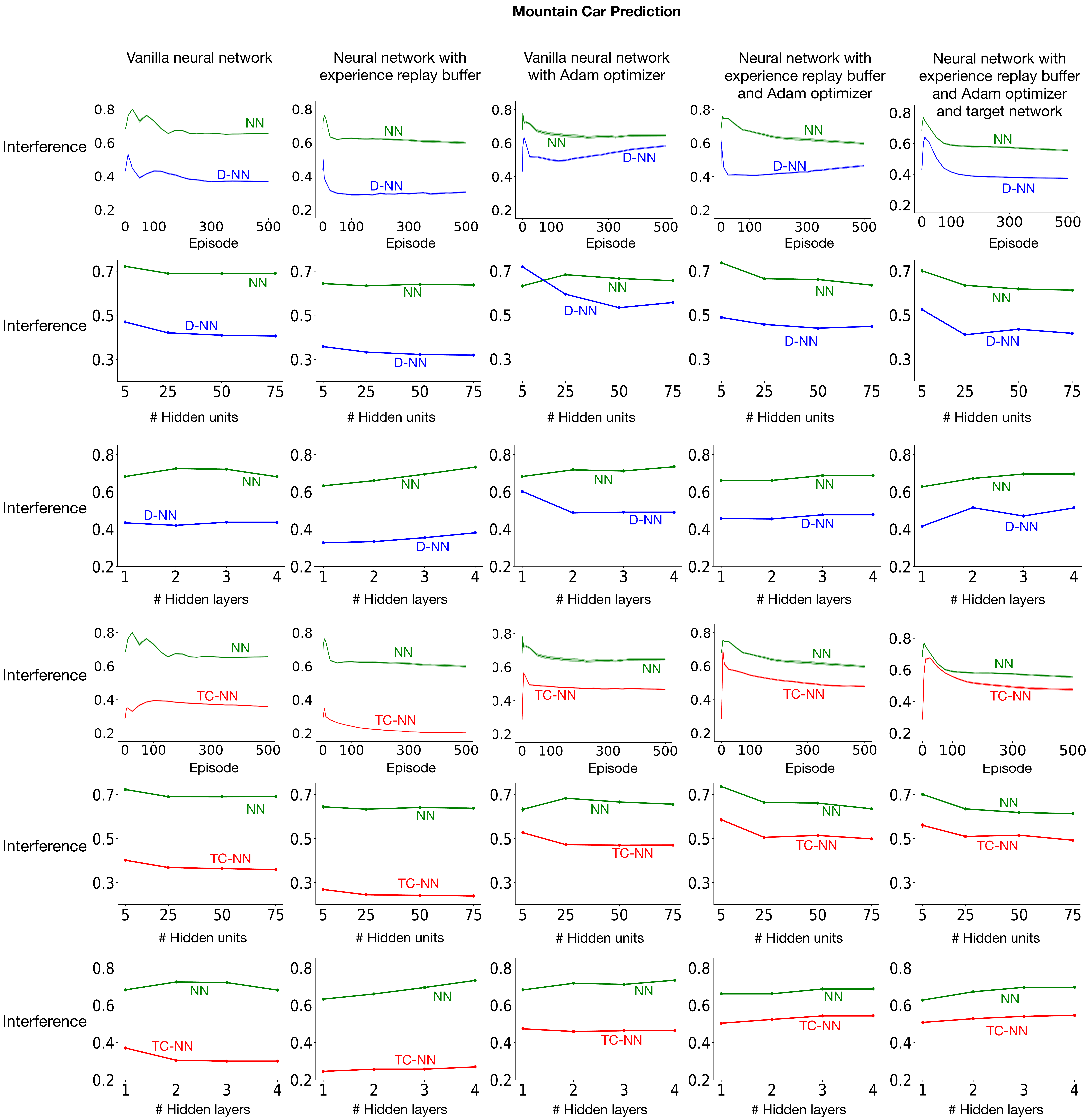}
      \caption{Interference over time (first and fourth row), interference over the number of hidden units (second and fifth row), and interference over the number of hidden layers (third and sixth row) for discretized vs. raw and tile coding vs. raw are shown. Breaking the generalization in the input space reduced the interference.}
      \label{fig:interference}
\end{figure*}

The results in the previous section indicate that input preprocessing can improve learning speed, stability, and parameter sensitivity, but it is not clear how these improvements relate to interference in the network updates. One hypothesis is that preprocessing reduces interference. To test this hypothesis, we use a recently proposed measure of interference (Liu, 2019). The measure is only applicable to prediction tasks and so the results in this section are only suggestive of the possible relationship between performance improvements due to input preprocessing and reduction in interference. 

Liu’s measure is fairly straightforward and is based on the Pairwise Interference (PI):
\begin{align}
 \text{PI}(S_i, S_j) = \frac{\nabla_\vecw\big[\hat{v}(\vecw, S_i)\big]\tr \nabla_\vecw\big[\hat{v}(\vecw, S_j)\big]}{||\nabla_\vecw\hat{v}(\vecw, S_i)||_2 \times ||\nabla_\vecw\hat{v}(\vecw, S_j)||_2}, \label{eq:interference}
\end{align}
\noindent which means if the two vectors in the numerator have a zero inner product, they are orthogonal and update made to $S_i$ does not affect the update from $S_j$. This  is similar to the neural tangent kernel that has been previously used for studying generalization in deep reinforcement learning (Achaim, Knight and Abbeel, 2019), scaled by the norm. The measure is also similar to the one proposed by French (1999) for batch supervised learning. 

We measured the pairwise interference (Equation~\ref{eq:interference}) for all pairs of samples in the dataset $\mathcal{D}$ where $\mathcal{D}$ is the same as the one used to compute $\text{R}\overline{\text{VE}}$. See Equation~\ref{eq:RVE}. We averaged the pairwise interference over all samples and then averaged the results over 30 runs. We repeated this process as learning progressed, specifically, before the learning started, after initializing the neural network, and then after episodes 1, 5, 10, 25, and every 25 episodes afterwards until the end of learning (500 episodes in this case). The interference for each method was measured for the parameter setting that produced the best learning curve as explained in Section \ref{subsct:OverallPerformance}.

To sanity check this Liu’s measure, we can compare the interference of a learning system we expect to have high interference with one we expect to have low interference and see if the measure agrees. Let’s consider the simple feed-forward NN with SGD shown in Figure \ref{fig:interference} (first row, leftmost subplot). We can compare this with the measured interference of a NN with ER, Adam and target networks (Figure \ref{fig:interference}, first row, rightmost subplot). We can see the simple NN results in higher measured interference than the NN with target networks, ER, and Adam.. 

Rows 1 and 4 of Figure~\ref{fig:interference}, compare the interference of different methods over episodes. The parameters for these interference measurements were the same as the ones used to produce learning curves in Figure~\ref{fig:learning_curve}. Discretization and tile coding both reduced the interference compared to when raw input is used.

% \subsection{Interference over the number of hidden layers}
% \label{subsct:InterferenceOverTheNumberOfLayers}

We compared the interference of different methods with different numbers of hidden layers in the network. We fixed the number of hidden units at each layer to 25. To calculate a single number that represents the interference of each setting, we measured the interference over time, as discussed before for each run. We then averaged the results over time to get a single interference measurement for each run. We then computed the average and standard deviation of the resulting numbers as a measure of interference for each setting. Rows 3 and 6 of Figure~\ref{fig:interference}, show the interference over different number of hidden layers, 1 layer to 4 layers. In almost all cases, the interference was reduced when using tile coding or discretization. The interference was rather insensitive to number of hidden layers.

% \subsection{Interference over the number of hidden units}
% \label{subsct:Interference}

We also measured the interference over different number of hidden units. We fixed the number of hidden layers to 1, and changed the number of hidden units between 5, 10, 25, 50, 75 units. The process was the same as what we used to compute interference for increasing number of hidden layers, described before.

Figure~\ref{fig:interference}, rows 2 and 5, shows that the interference generally decreased with increasing number of hidden units. This is possibly because larger networks have more capacity to reduce the impact of input generalization.

\section{Conclusion and future work}
\label{sct:ConclusionAndFutureWork}

Scaling up reinforcement learning requires function approximation, but it is still not clear how to best learn the representation online. Modern NN systems work well in control tasks that would be nearly impossible with hand-designed static features. The NN learning systems used today employ a host of algorithmic tricks to achieve stable learning, but these systems are still slow to train and can exhibit significant variability in performance. Some of these limitations can potentially be explained by catastrophic interference in the network updates, which has been relatively under-explored in online reinforcement learning. In this paper, we take some first steps toward understanding how online NN training and interference relate. We find that simply re-mapping the input observations to a high-dimensional space that breaks the input generalization, improves learning speed, and parameter sensitivity across several different architectures on two different control tasks. We also show empirically that this preprocessing can reduce interference in prediction tasks. More practically, we provide a simple approach to NN training that is easy to implement, requires little additional computation, and can significantly improve performance.  

One possible next step is to study the performance of these methods in higher dimensional spaces. We used simple classic reinforcement learning tasks in this paper so we could perform extensive hyper-parameter analysis and average over many runs. Nevertheless, the discretization methods proposed here is highly scalable (input dimension grows only linearly in the number of bins), but it remains unclear how effective it will be in higher dimensional spaces. Another possible next step is to investigate other preprocessing schemes and a greater variety of base learning systems. Here we explored discretization and tile coding because they are simple and their impact on the learning system is easy to understand. Previous work demonstrated that simple strategies like tile coding can be applied to rather high-dimensional spaces on mobile robots by tiling the dimensions independently (Stone and Sutton, 2001; Modayil et al., 2014; Rafiee et al., 2019); tile coding based preprocessing may work well in larger problems. There are likely better preprocessing strategies, that may even result in performance competitive with more complex deep reinforcement learning systems. 

There is more work to be done characterizing and correlating poor performance in NN learning with interference. We have just begun to understand how sparse representations (Liu et al., 2019), training to reduce interference (Javed and White, 2019), and input preprocessing---our work---improves NN training. Perhaps we need better optimizers, or to dynamically grow the network over time. It is still unclear if interference is worse due to changes in the policy, or if there are additional complications for algorithms that use bootstrap targets like TD. Whatever the solution, we need to clearly define and measure interference before we can hope to mitigate it.

\section{Implementation Details}
\label{sct:ImplementationDetails}
In this section, we describe our most important design decisions and parameter settings to aid reproducibility. We used Pytorch version 1.3.0 for training the neural networks (Paszke et al., 2017). All weights in the neural networks were initialized using Xavier uniform initialization (Glorot and Bengio, 2010). All NN used ReLU activation in their hidden layers and were linear in the output. We tested several values of $\beta_1$ and $\beta_2$ to achieve good performance with Adam: all possible combinations of $\beta_1\in\{0.9, 0.99, 0.999\}$ and $\beta_2\in\{0.9, 0.99, 0.999, 0.9999\}$. We used a replay buffer size of 2000. The target networks were updated every 100 steps (this value was determined through trial-and-error parameter tuning)\footnote{We also tried updating the target network every 50 or 10 steps instead of 100, the results of which were similar to the results presented in the paper.}. We used a mini-batch size of 32 for all network updates. In control experiments, we used Sarsa(0) with an $\epsilon$-greedy behavior and $\epsilon = 0.1$.

There are a few details that were problem specific. In Mountain Car, the value function was approximated with a NN with one hidden layer and 50 hidden units, and in Acrobot the hidden layer contained 100 hidden units. In control, the network outputted three values, one for each action.

Input preprocessing was fairly generic and similar across both problems. The input to the networks was normalized between -1 and 1 when no preprocessing was used. For discretization-based preprocessing, each input dimension was discretized independently and passed to the network as concatenated one-hot vectors. In Mountain Car, the inputs were discretized into 20 bins per dimension, and in Acrobot we used 32 bins per dimension. For tile coding based preprocessing, we tiled each input dimension together producing a binary vector as input to the NN. We used 8 tilings, with a tile width of 1/4 for each dimension: \(4 \times 4\) tiles for Mountain Car yielding 128 length binary vectors and \(4 \times 4 \times 4 \times 4\) tiles for Acrobot hashed into 512 length binary vectors. We tested both joint tiling the dimensions and independent tilings. The results reported in Section 5 used joint tilings. We used the freely available Python tile coding software\footnote{http://incompleteideas.net/tiles/tiles3.html.}.

% \usepackage{multirow}
\begin{table}[]
\caption{Methods and parameters used for different tasks.}
\footnotesize
\begin{tabular}{|c|c|c|c|c|}
\hline
\multicolumn{2}{|c|}{Methods} & Adam $\beta_1$ \& $\beta_2$ & $\alpha$ & Additional \\ \hline
\multirow{5}{*}{NN} & SGD & --- & \multirow{15}{*}{\begin{tabular}[c]{@{}c@{}}Mountain car\\ prediction\\ $2^{-c}, c \in$\\ \{3, 4, ..., 18\}\\ \\ \\ \\ Mountain car\\ control\\ $2^{-c}, c \in$\\ \{1, 2, ..., 18\}\\ \\ \\ \\ Acrobot\\ $2^{-c}, c \in$\\ \{5, 6, ... , 18\}\end{tabular}} & \multirow{15}{*}{\begin{tabular}[c]{@{}c@{}}Target net \\ update frequency $\in$ \\ \{10, 50, 100\}.\\ Set to 100 in results.\\ \\ TC-NN:\\ Mountain car:\\ $4\times 4$ tiles,\\ 8 tilings.\\ Acrobot:\\ $4\times 4\times 4$ tiles,\\ 8 tilings.\\ \\ D-NN:\\ Mountain car:\\ 20 bins/dimension.\\ Acrobot: \\ 32 bins/dimension\end{tabular}} \\ \cline{2-3}
 & SGD+ER & --- &  &  \\ \cline{2-3}
 & Adam & \multirow{3}{*}{\begin{tabular}[c]{@{}c@{}}$\beta_1 \in$ \{0.999,\\ 0.99, 0.9\} and\\ $\beta_2 \in$ \{0.9999,\\ 0.999, 0.99, 0.9\}\end{tabular}} &  &  \\ \cline{2-2}
 & Adam+ER &  &  &  \\ \cline{2-2}
 & \begin{tabular}[c]{@{}c@{}}Adam+ER\\ +TN\end{tabular} &  &  &  \\ \cline{1-3}
\multirow{5}{*}{\begin{tabular}[c]{@{}c@{}}D-\\ NN\end{tabular}} & SGD & --- &  &  \\ \cline{2-3}
 & SGD+ER & --- &  &  \\ \cline{2-3}
 & Adam & \multirow{3}{*}{\begin{tabular}[c]{@{}c@{}}$\beta_1 \in$ \{0.999,\\ 0.99, 0.9\} and\\ $\beta_2 \in$ \{0.9999,\\ 0.999, 0.99, 0.9\}\end{tabular}} &  &  \\ \cline{2-2}
 & Adam+ER &  &  &  \\ \cline{2-2}
 & \begin{tabular}[c]{@{}c@{}}Adam+ER\\ +TN\end{tabular} &  &  &  \\ \cline{1-3}
\multirow{5}{*}{\begin{tabular}[c]{@{}c@{}}TC-\\ NN\end{tabular}} & SGD & --- &  &  \\ \cline{2-3}
 & SGD+ER & --- &  &  \\ \cline{2-3}
 & Adam & \multirow{3}{*}{\begin{tabular}[c]{@{}c@{}}$\beta_1 \in$ \{0.999,\\ 0.99, 0.9\} and\\ $\beta_2 \in$ \{0.9999,\\ 0.999, 0.99, 0.9\}\end{tabular}} &  &  \\ \cline{2-2}
 & Adam+ER &  &  &  \\ \cline{2-2}
 & \begin{tabular}[c]{@{}c@{}}Adam+ER\\ +TN\end{tabular} &  &  &  \\ \hline
\end{tabular}
\label{tbl:parameter_list}
\end{table}

\section*{ACKNOWLEDGMENTS}
The authors thank Richard Sutton,Vincent Liu, and Shivam Garg for insights and helpful comments. Andrew Patterson helped proofreading the paper. The authors gratefully acknowledge funding from JPMorgan Chase \& Co., Google DeepMind, Natural Sciences and Engineering Research Council of Canada (NSERC), and Alberta Innovates—Technology Futures (AITF).

\clearpage
\section*{References}
\small

\begin{list}{}{%
\setlength{\topsep}{0pt}%
\setlength{\leftmargin}{0.2in}%
\setlength{\listparindent}{-0.2in}%
\setlength{\itemindent}{-0.2in}%
\setlength{\parsep}{\parskip}%
}%

\item[]
[1] Achiam, J., Knight, E., and Abbeel, P. (2019). Towards Characterizing Divergence in Deep Q-Learning. arXiv:1903.08894.

\item[]
[2] Albus, J. S. (1975). Data storage in the Cerebellar Model Articulation Controller (CMAC). \textit{Journal of Dynamic Systems, Measurement and Control}, 97(3):228–233.

\item[]
[3] Albus, J. S. (1981). \textit{Brains, Behavior, and Robotics.} Peterborough, NH.

\item[]
[4] Brockman, G.; Cheung, V.; Pettersson, L.; Schneider, J.; Schulman, J.; Tang, J.; and Zaremba, W. (2016). Openai gym. arXiv:1606.01540.

\item[]
[5] Duan, Y., Chen, X., Houthooft, R., Schulman, J., Abbeel, P. (2016, June). Benchmarking deep reinforcement learning for continuous control. In \textit{International Conference on Machine Learning} (pp. 1329-1338).

\item[]
[6] French, R. M. 1999. Catastrophic forgetting in connectionist net- works: Causes, consequences and solutions. \textit{Trends in Cognitive Sciences,} \\3(4):128–135.

\item[]
[7] Glorot, X., and Bengio, Y. (2010). Understanding the difficulty of training deep feedforward neural networks. In \textit{Proceedings of the thirteenth international conference on artificial intelligence and statistics}.

\item[]
[8] Goodfellow I. J., Mirza M, Xiao D, Courville A, Bengio Y. An empirical investigation of catastrophic forgetting in gradient-based neural networks. arXiv:1312.6211. 2013 Dec 21.

\item[]
[9] Goodrich, B. F. (2015). Neuron clustering for mitigating catastrophic forgetting in supervised and reinforcement learning.

\item[]
[10] Henderson, P., Islam, R., Bachman, P., Pineau, J., Precup, D. and Meger, D., 2018, April. Deep reinforcement learning that matters. In \textit{Thirty-Second AAAI Conference on Artificial Intelligence.}

\item[]
[11] Jacobsen, A., Schlegel, M., Linke, C., Degris, T., White, A., and White, M. (2019). Meta-descent for Online, Continual Prediction. In \textit{AAAI Conference on Artificial Intelligence.}

\item[]
[12] Javed, K., and White, M. (2019). Meta-learning representations for continual learning. In \textit{Advances in Neural Information Processing Systems} (pp. 1818-1828).

\item[]
[13] Kingma, D.P. and Ba, J., (2014). Adam: A method for stochastic optimization. arXiv:1412.6980.

\item[]
[14] Kirkpatrick, J., Pascanu, R., Rabinowitz, N., Veness, J., Desjardins, G., Rusu, A. A.,  Milan, K., Quan, J., Ramalho, T., Grabska-Barwinska, A. and Hassabis, D. (2017). Overcoming catastrophic forgetting in neural networks. \textit{Proceedings of the national academy of sciences}, 114(13), 3521-3526.

\item[]
[15] Lin, L. J. (1992). Self-improving reactive agents based on reinforcement learning, planning and teaching. \textit{Machine learning}, 8(3-4), 293-321.

\item[]
[16] Liu. V. (2019). \textit{Sparse Representation Neural Networks for Online Reinforcement Learning}. M.Sc. thesis, University of Alberta, Edmonton, Canada.

\item[]
[17] Liu, V. Kumaraswamy, R., Le, L., and White, M. (2019). The utility of sparse representations for control in reinforcement learning. In \textit{Proceedings of the AAAI Conference on Artificial Intelligence} (Vol. 33, pp. 4384-4391).

\item[]
[18] Mahmood, A. R., Sutton, R. S. (2013). Representation Search through Generate and Test. In \textit{Proceedings of the AAAI Workshop on Learning Rich Representations from Low-Level Sensors}, Bellevue, WA, USA.

\item[]
[19] McCloskey, M., and Cohen, N. J. (1989). Catastrophic interference in connectionist networks: The sequential learning problem. In \textit{Psychology of learning and motivation} (Vol. 24, pp. 109-165). Academic Press.

\item[]
[20] Minsky, M., and Papert, S. A. (2017). \textit{Perceptrons: An introduction to computational geometry.} MIT press.

\vfill\eject

\item[]
[21] Mnih, V., Kavukcuoglu, K., Silver, D., Rusu, A. A., Veness, J., Bellemare, M. G., Graves, A., Riedmiller, M., Fidjeland, A. K., Ostrovski, G., Petersen, S., Beattie, C., Sadik, A., Antonoglou, I., King, H., Kumaran, D., Wierstra, D., Legg, S., Hassabis, D. (2015). Human level control through deep reinforcement learning. \textit{Nature, 518}(7540):529–533.

\item[]
[22] Modayil, J., White, A., and Sutton, R. S. (2014). Multi-timescale nexting in a reinforcement learning robot. \textit{Adaptive Behavior}, 22(2), 146-160.

\item[]
[23] Moore, A. W. (1991). Variable resolution dynamic programming: Efficiently learning action maps in multivariate real-valued state-spaces. In \textit{Machine Learning Proceedings 1991} (pp. 333-337). Morgan Kaufmann.

\item[]
[24] Parisotto, E. and Salakhutdinov, R. (2017). Neural map: Structured memory for deep reinforcement learning. arXiv:1702.08360.

\item[]
[25] Puterman, M. L. (2014). Markov Decision Processes.: Discrete Stochastic Dynamic Programming. John Wiley and Sons.

\item[]
[26] Rafiee, B., Ghiassian, S., White, A., and Sutton, R. S. (2019, May). Prediction in Intelligence: An Empirical Comparison of Off-policy Algorithms on Robots. In \textit{Proceedings of the 18th International Conference on Autonomous Agents and MultiAgent Systems} (pp. 332-340). International Foundation for Autonomous Agents and Multiagent Systems.

\item[]
[27] Rajeswaran, A., Lowrey, K., Todorov, E. V., and Kakade, S. M. (2017). Towards generalization and simplicity in continuous control. In \textit{Advances in Neural Information Processing Systems} (pp. 6550-6561).

\item[]
[28] Riedmiller, M., Hafner, R., Lampe, T., Neunert, M., Degrave, J., Van de Wiele, T., Mnih, V., Heess, N., Springenberg, J. T. and Springenberg, J. T. (2018). Learning by playing-solving sparse reward tasks from scratch. arXiv:1802.10567.

\item[]
[29] Rummery, G. A., and Niranjan, M. (1994). \textit{On-line Q-learning using connectionist systems} (Vol. 37). Cambridge, England: University of Cambridge, Department of Engineering.

\item[]
[30] Silver, D. (2009). Reinforcement learning and simulation-based search in computer Go.

\item[]
[31] Stone, P., Kuhlmann, G., Taylor, M. E., and Liu, Y. (2005). Keepaway soccer: From machine learning testbed to benchmark. In \textit{Robot Soccer World Cup} (pp. 93-105). Springer, Berlin, Heidelberg.

\item[]
[32] Stone, P., and Sutton, R. S. (2001). Scaling reinforcement learning toward RoboCup soccer. In \textit{Icml} (Vol. 1, pp. 537-544).

\item[]
[33] Sturtevant, N. R., and White, A. M. (2006, May). Feature construction for reinforcement learning in hearts. In \textit{International Conference on Computers and Games} (pp. 122-134). Springer, Berlin, Heidelberg.

\item[]
[34] Sutton, R. S. (1996). Generalization in reinforcement learning: Successful examples using sparse coarse coding. In \textit{Advances in neural information processing systems} (pp. 1038-1044).

\item[]
[35] Sutton, R. S. (1988). Learning to predict by the methods of temporal differences. \textit{Machine learning}, 3(1), 9-44.

\item[]
[36] Sutton, R. S., and Barto, A. G. (2018). Introduction to reinforcement learning. Cambridge, MA: MIT Press, 2nd edition.

\item[]
[37] Sutton, R. S., and Whitehead, S. D. (1993). Online learning with random representations. In \textit{Proceedings of the Tenth International Conference on Machine Learning} (pp. 314-321).

\item[]
[38] Tesauro, G. (1995). Temporal difference learning and TD-Gammon. \textit{Communications of the ACM}, 38(3), 58-68.

\item[]
[39] Vinyals, O., Babuschkin, I., Chung, J., Mathieu, M., Jaderberg, M., Czarnecki, W. M., Dudzik, A., Huang, A., Georgiev, P., Powell, R. and Ewalds, T. (2019). AlphaStar: Mastering the real-time strategy game StarCraft II. \textit{DeepMind Blog}.

\item[]
[40] Zhang, S., and Sutton, R. S. (2017). A deeper look at experience replay. arXiv:1702.01275.

\end{list}

\onecolumn
\appendix

\section{Two sample t-test results}

% \usepackage{multirow}
\begin{table}[h!]
\caption{P-values of the AUC of the learning curves. The difference between the AUC of the learning curves is significant.}
\begin{tabular}{|c|c|c|c|c|c|c|}
\hline
\multicolumn{2}{|c|}{} & Vanilla neural network & \begin{tabular}[c]{@{}c@{}}Neural netowrk\\ with experience\\ replay buffer\end{tabular} & \begin{tabular}[c]{@{}c@{}}Vanilla neural\\ network with\\ Adam optimizer\end{tabular} & \begin{tabular}[c]{@{}c@{}}Neural network\\ with experience\\ replay buffer and\\ Adam optimizer\end{tabular} & \begin{tabular}[c]{@{}c@{}}Neural network\\ with experience\\ replay buffer and\\ Adam optimizer\\ and target network\end{tabular} \\ \hline
\multirow{3}{*}{\begin{tabular}[c]{@{}c@{}}Discretized input\\ versus\\ raw input\\ (D-NN\\ vs. NN)\end{tabular}} & \begin{tabular}[c]{@{}c@{}}Mountain Car\\ Prediction\end{tabular} & $2.58 \times 10^{-50}$ & $8.59 \times 10^{-9}$ & $1.19 \times 10^{-36}$ & $6.33 \times 10^{-11}$ & $9.01 \times 10^{-19}$ \\ \cline{2-7} 
 & \begin{tabular}[c]{@{}c@{}}Mountain Car\\ Control\end{tabular} & $1.61 \times 10^{-13}$ & $2.07 \times 10^{-19}$ & $6.94 \times 10^{-12}$ & $2.57 \times 10^{-18}$ & $1.76 \times 10^{-2}$ \\ \cline{2-7} 
 & \begin{tabular}[c]{@{}c@{}}Acrobot\\ \vspace{1mm}\end{tabular} & $7.00 \times 10^{-16}$ & $3.47 \times 10^{-7}$ & $3.42 \times 10^{-10}$ & $2.95 \times 10^{-2}$ & $1.09 \times 10^{-5}$ \\ \hline
\multirow{3}{*}{\begin{tabular}[c]{@{}c@{}}Tile coded input\\ versus\\ raw input\\ (TC-NN\\ vs. NN)\end{tabular}} & \begin{tabular}[c]{@{}c@{}}Mountain Car\\ Prediction\end{tabular} & $9.97 \times 10^{-56}$ & $1.12 \times 10^{-22}$ & $2.29 \times 10^{-40}$ & $8.75 \times 10^{-12}$ & $8.31 \times 10^{-28}$ \\ \cline{2-7} 
 & \begin{tabular}[c]{@{}c@{}}Mountain Car\\ Control\end{tabular} & $3.35 \times 10^{-18}$ & $5.52 \times 10^{-21}$ & $3.76 \times 10^{-14}$ & $1.26 \times 10^{-18}$ & $2.25 \times 10^{-7}$ \\ \cline{2-7} 
 & \begin{tabular}[c]{@{}c@{}}Acrobot\\ \vspace{1mm}\end{tabular} & $1.49 \times 10^{-17}$ & $4.56 \times 10^{-8}$ & $1.83 \times 10^{-22}$ & $1.16 \times 10^{-2}$ & $1.08 \times 10^{-10}$ \\ \hline
\end{tabular}
\label{tab:p-values}
\end{table}

We performed a two sample t-test on the area under the learning curves reported in Figure~\ref{fig:learning_curve}. Table~\ref{tab:p-values} summarizes the p-values of the test. Each cell in the table reports the p-value from the two sample t-test comparing the 30 AUCs (one for each run) of the corresponding learning curve in Figure~\ref{fig:learning_curve}. The significance test rejects the null hypothesis that the area under the learning curves come from normal distribution with equal means at the 5\% level.

\end{document}